\documentclass[12pt]{article}
\usepackage{graphicx}
\usepackage[margin=1in]{geometry}
\usepackage{setspace}
\usepackage{wrapfig}
\usepackage{lmodern}
\usepackage{amsmath, amssymb}
\usepackage{float}
\usepackage{caption}
\usepackage{subcaption}
\usepackage{url} 
\title{Analyzing the Structure of Handwritten Digits: 
A Comparative Study of PCA, Factor Analysis, and UMAP}

\author{
Jyotiraditya Gupta\\
\small Department of Statistical Sciences\\
\small Department of Computer Science\\
\small University of Toronto
}

\date{}

\begin{document}

\maketitle
\begin{abstract}
Handwritten digit images lie in a high-dimensional pixel space but exhibit strong geometric
and statistical structure. This paper investigates the latent organization of handwritten
digits in the MNIST dataset using three complementary dimensionality reduction techniques:
Principal Component Analysis (PCA), Factor Analysis (FA), and Uniform Manifold Approximation
and Projection (UMAP). Rather than focusing on classification accuracy, we study how each
method characterizes intrinsic dimensionality, shared variation, and nonlinear geometry.
PCA reveals dominant global variance directions and enables high-fidelity reconstructions
using a small number of components. FA decomposes digits into interpretable latent handwriting
primitives corresponding to strokes, loops, and symmetry. UMAP uncovers nonlinear manifolds
that reflect smooth stylistic transitions between digit classes. Together, these results
demonstrate that handwritten digits occupy a structured low-dimensional manifold and that
different statistical frameworks expose complementary aspects of this structure.
\end{abstract}

\noindent\textbf{Reproducibility:} Experiments were conducted in Python using scikit-learn for PCA/FA and umap-learn for UMAP; key hyperparameters are reported in the text.

\section{Introduction}

The goal of this paper is to investigate the underlying structure of handwritten digits in the MNIST dataset using classical and modern statistical dimensionality reduction techniques.

Although MNIST is frequently used in machine learning for classification, we focus on understanding how high-dimensional image data can be compressed,
visualized, and interpreted using multivariate statistical methods.

Each image in the MNIST dataset contains rich geometric structure such as loops, edges, and symmetry, residing in a high-dimensional space (784 dimensions). This makes MNIST an ideal dataset for studying questions such as:
\begin{itemize}
    \item How many dimensions are needed to represent handwritten digits without substantial information loss?
    \item How does the number of retained components affect reconstruction quality, and what does this imply about intrinsic dimensionality of MNIST digits?
     \item How do PCA and FA differ in the latent structure they extract from handwritten digits, and what does this reveal about the common and total variation in handwriting styles?
    \item Does a nonlinear method like UMAP uncover digit-level separability or geometric patterns that linear methods fail to capture?
  
\end{itemize}

\subsection{Background and Scope}

The MNIST dataset is well suited for studying high-dimensional statistical structure because it combines rich geometric variation with direct visual interpretability. MNIST is high-dimensional (784 features per image) yet structured in a way that makes dimensionality-reduction techniques both meaningful and visually interpretable. Each image can be reshaped back into a 28×28 grid, allowing principal components, factor loadings, and UMAP embeddings to be directly visualized as eigen-digits or geometric patterns.

\section{Data Description}

\subsection{Data overview}
The dataset used in this study is the MNIST handwritten digits dataset, consisting of
5,000 grayscale images of digits from 0 to 9. Each image is of size $28 \times 28$ pixels, resulting in a 784-dimensional representation when flattened. Pixel intensities range
from 0 (black) to 255 (white). 
\subsection{Preprocessing}

Data exploration confirmed that the dataset 
contains \textbf{no missing values}, and all pixel intensities lie in the integer range 
$[0, 255]$. The following preprocessing steps were performed:

\begin{itemize}
    \item \textbf{Flattening:}  
    Each image was reshaped from a $28\times 28$ grid into a 784-dimensional vector to 
    enable linear-algebra operations required for PCA, Factor Analysis, and UMAP.
    
    \item \textbf{Variance filtering for correlation analysis:}  
    When computing the pixel--pixel correlation matrix, pixels with zero variance were 
    removed, since constant-valued features cause undefined correlations.  
    This filtering step was used \emph{only} for the correlation heatmap, and does not 
    affect PCA, FA, or UMAP results.

        \item \textbf{No Standardization:}  
    Standardization was not used because all pixel features share the same 
    physical meaning and scale, standardizing each pixel to zero mean and unit variance would artificially amplify background noise (since most pixels are zero) and distort the image geometry.
\end{itemize}

No additional smoothing or dimensionality transformations were
performed.

\subsection{Exploratory Analysis and Summary Statistics}

We confirmed that all ten digit classes (0 to 9) have equal representations (500 images each). A median value of 0 (\textbf{Figure 1}) implies that over half of the total pixels are dark (black), which makes sense intuitively since each handwritten image will have more empty space (dark pixels) than actual text (bright pixels). More summary stats are displayed in the \textbf{Figure 1}. 

\begin{figure}[h]
    \centering
    \includegraphics[width=0.25\textwidth]{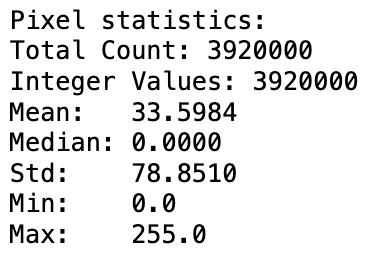}
    \caption{Basic Summary Statistics}
\end{figure}

We visualized several random digits to confirm the integrity of the data. Next, we ran a simple pixel correlation (\textbf{Figure 2}). The first few and the last few pixels seem to be filtered out, since we only plot the correlation for pixels which have a non-zero variance across the dataset. Nearby pixels seem to have a consistently high correlation, which is a sensible result given the nature of the data, where brighter pixels would likely have bright neighbouring pixels, and vice-versa for darker pixels.
\begin{figure}[h]
    \centering
    \includegraphics[width=0.65\textwidth]{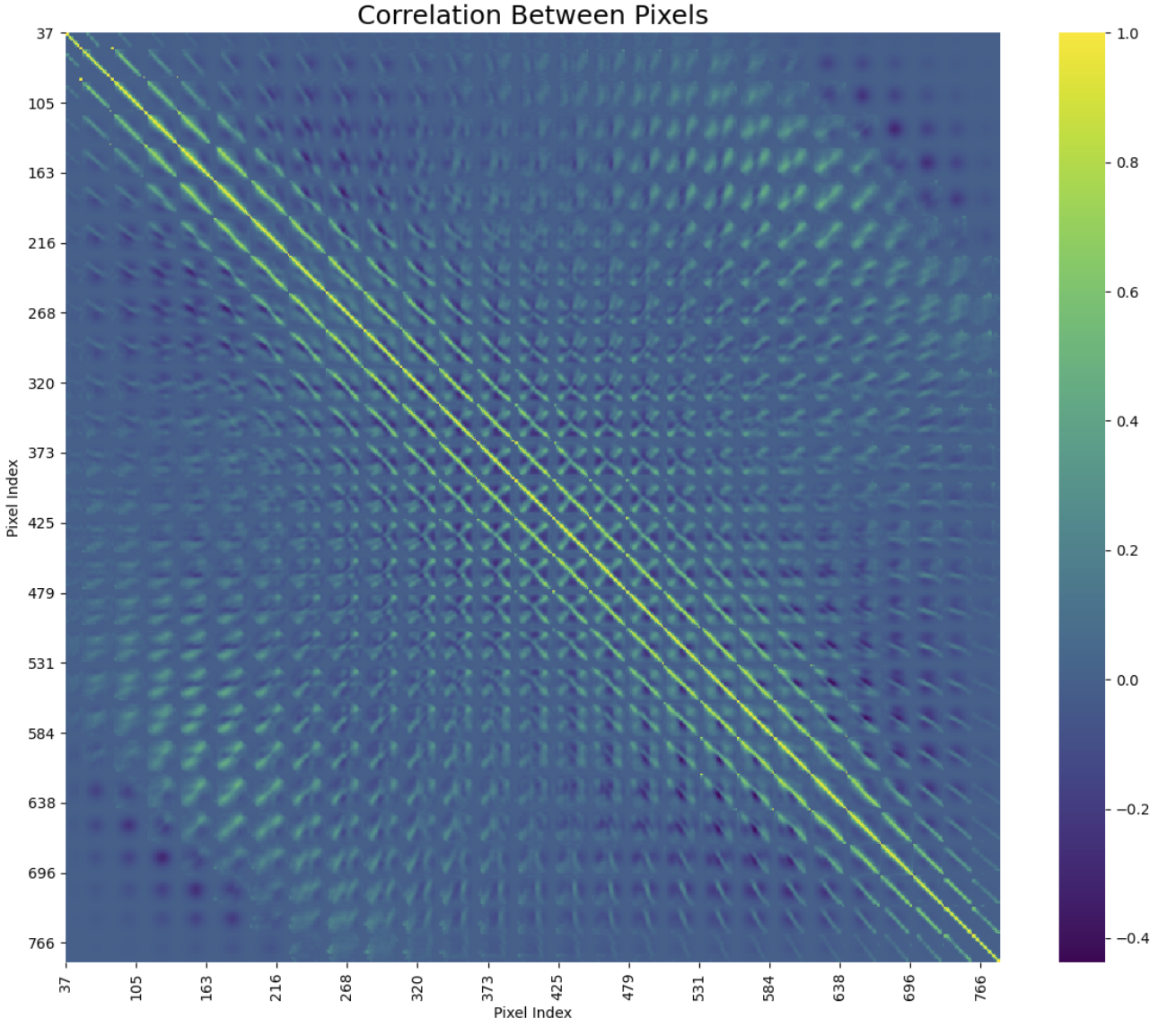}
    \caption{Basic Pixel-Correlation}
\end{figure}

\section{Methodology and Results}

In this section, we describe each statistical method used in the study. Each technique includes both a methodological overview and the empirical findings obtained from the dataset.

\subsection{Principal Component Analysis (PCA)}

\subsubsection*{Methodology}
Principal Component Analysis (PCA) was applied to reduce the 784-dimensional pixel space into a lower-dimensional representation. We used PCA because it identifies directions of maximal variance, produces interpretable eigenvectors (eigen-digits), and enables reconstruction of digits to quantify information loss.
We extracted the top 100 components and analyzed variance, embeddings, eigen-digits, and reconstruction quality.

\subsubsection*{Results}

\paragraph{Explained Variance.}
\begin{figure}[H]
    \centering
    \includegraphics[width=1.0\textwidth]{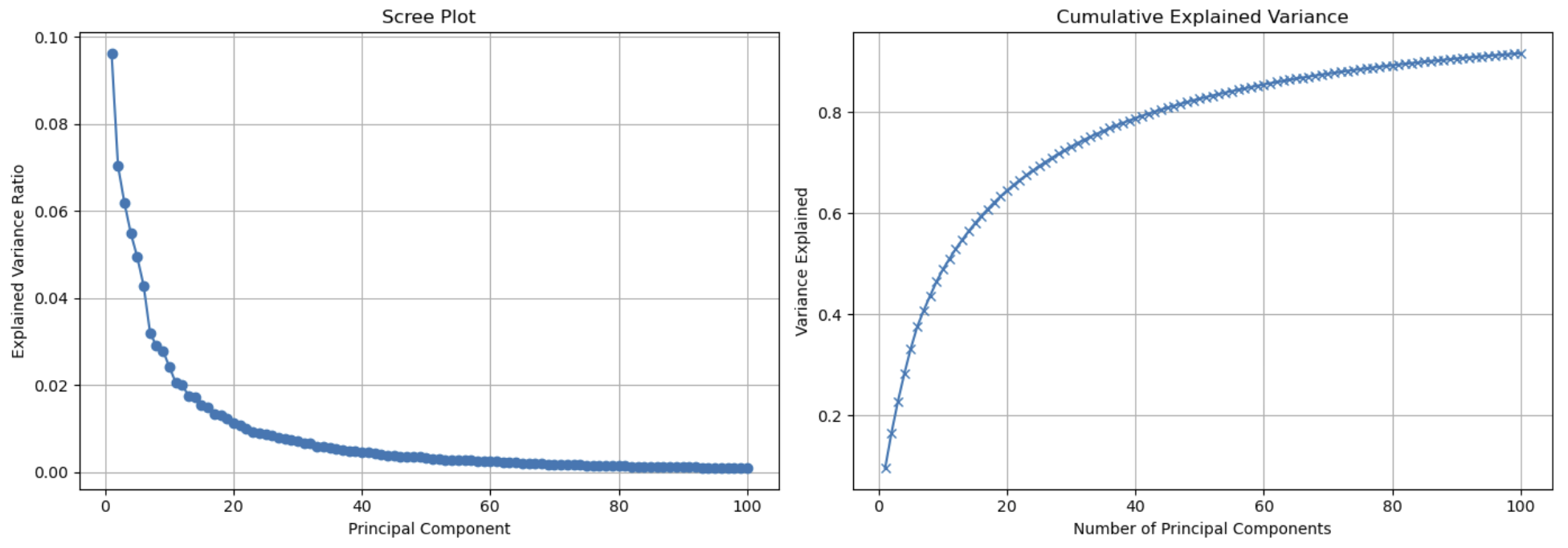}
    \caption{Variance across PCA components.}
\end{figure}
The cumulative explained variance curve (\textbf{Figure 3}) illustrates how much of the total variability 
in the dataset is captured as additional principal components are included. 
The scree plot exhibits a clear ``elbow,'' indicating that the first few components 
account for the majority of meaningful variation in the handwritten digits.

The first 10 principal components capture approximately 50\% of the total variance, 
revealing that pixel intensities are highly correlated and that only a small number 
of directions in feature space contain most of the structural information. 
Top 50 components explain over 80\%, of the variance. The plateau after around 50-60 components suggests that although each MNIST image resides 
in a 784-dimensional space, its intrinsic dimensionality could be far lower.

Most pixels contribute redundant information, background pixels contain almost no 
variance, and changes in handwriting shape tend to affect groups of pixels 
in coordinated ways. Overall, the variance profile strongly supports the 
use of dimensionality reduction for visualization, reconstruction, and further 
multivariate analysis.

\paragraph{PC Eigen-digits Analysis:}

\begin{figure}[H]
    \centering
    \includegraphics[width=1.0\textwidth]{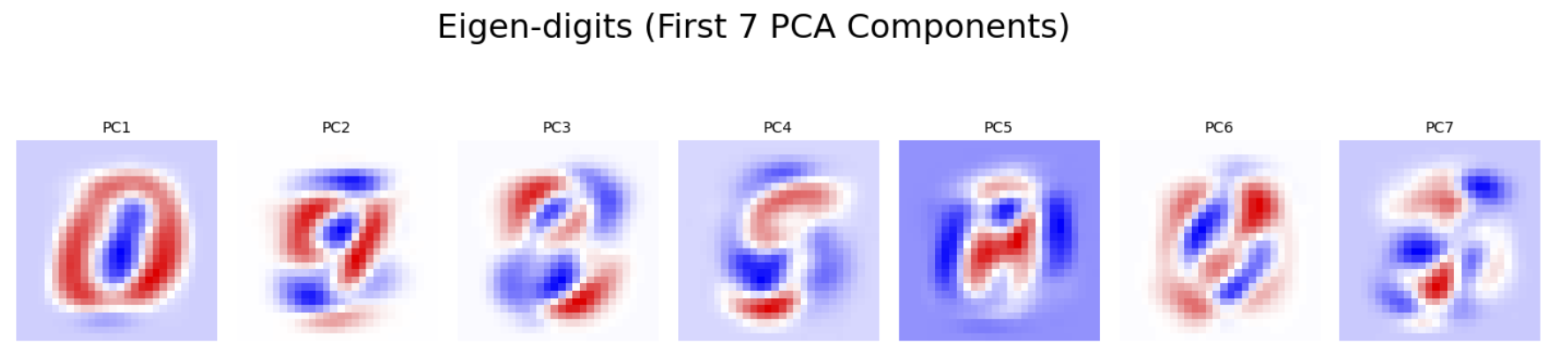}
    \caption{First seven PCA eigen-digits.}
\end{figure}

\begin{figure}[H]
    \centering
    \includegraphics[width=1.0\textwidth]{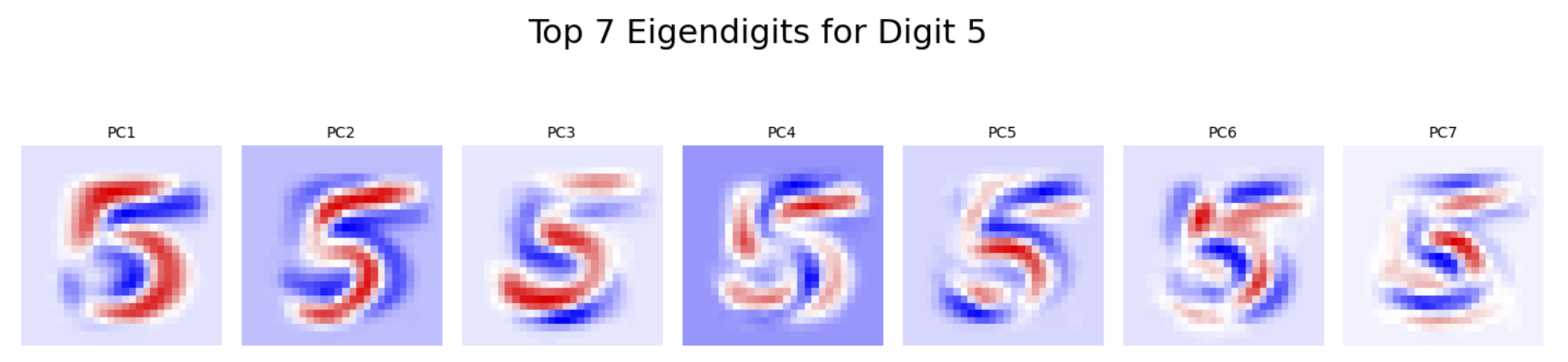}
    \caption{First seven PCA eigen-digits for the digit "5".}
\end{figure}
The first few PCA components in \textbf{Figure 4} reveal several sources of variation in handwritten digits. We choose only the top 7 PCs since the Scree Plot in \textbf{Figure 3} seems to have very slow degradation after the 7th component. The color theme has red for positive loadings and blue for negative loadings. This will be consistent throughout the analysis (for FA in the next section).
PC1 seems to capture global stroke thickness, while PC2 tries to capture nuances in slants and curves in the digits. PC3 seems to capture diagonal structure. Principal Components 4 to 7 capture spatial structures like top-bottom and left-right symmetry, loops and central core (which can be dark for digits like "0" and light for digits like "1").
We arbitrarily chose the digit “5” to analyze, and its eigendigits in \textbf{Figure 5} highlight consistent stylistic variations like the stroke thickness (PC1), slant of the top bar (PC2), curvature in the lower loop (PC3), and left-right symmetry (PC4).

\begin{figure}[H]
    \centering
    \includegraphics[width=0.7\textwidth]{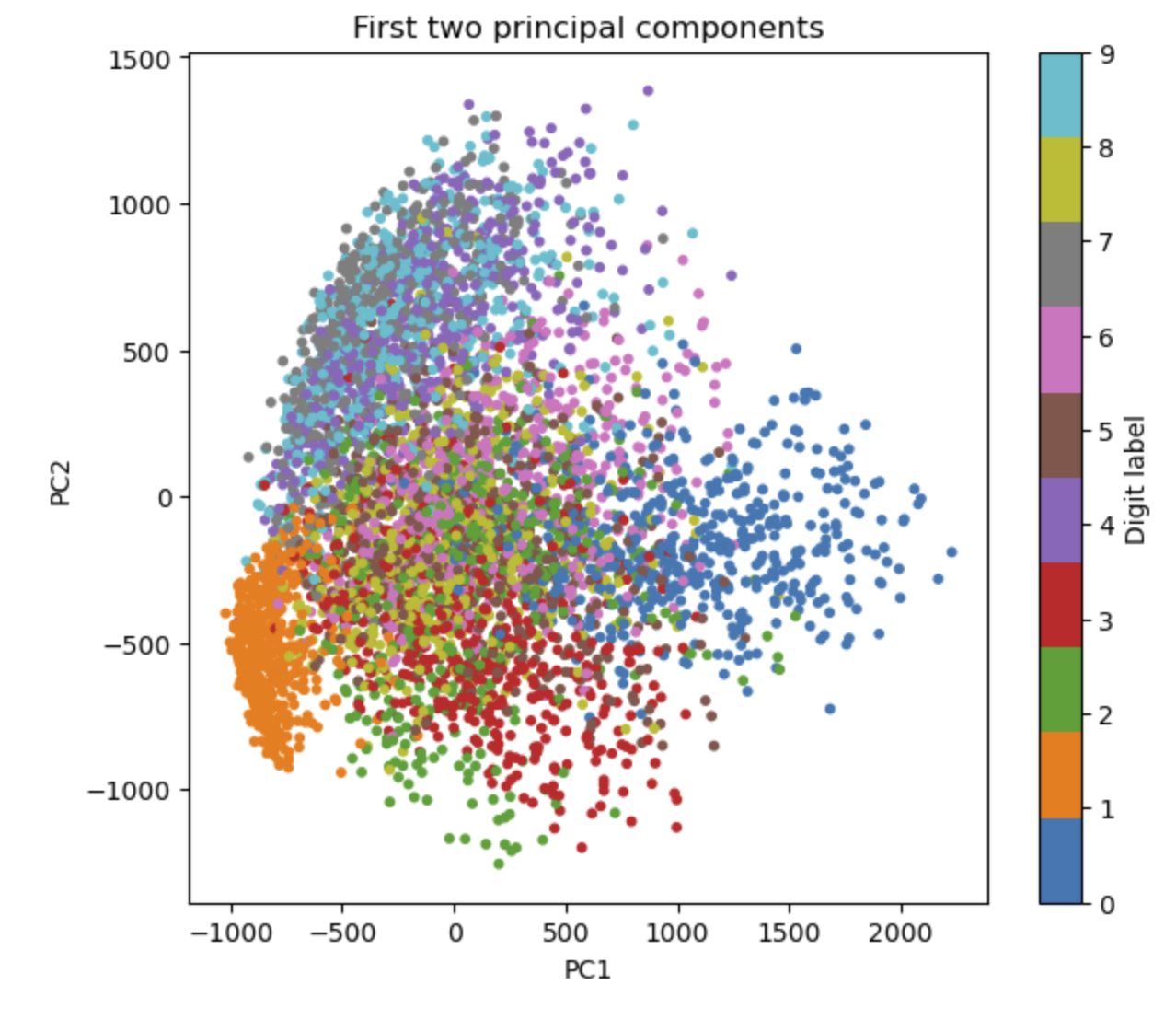}
    \caption{PCA embedding using the first two principal components.}
\end{figure}

The PCA embedding (\textbf{Figure 6}) shows that the first two principal components capture meaningful variation in handwriting despite PCA being a linear technique and the MNIST manifold being highly nonlinear. 

Digit "1" is well separated due to its uniquely simple, straight geometry. Moreover, it generally has a highly negative PC1 value, which aligns with the eigen-digit for PC1 from \textbf{Figure 4}, since digit "1" is mostly written in the centre, with minimal bright pixels on the boundary where PC1 has maximal variance. Digit "0" is also largely well separated, and naturally has the highest PC1 values since it is best correlated with the variance PC1 explains (bright strokes of overall thickness).  

However, digits with similar overall structure and curvature have significant overlap, for example "3", "5" and "8", as PCA cannot distinguish nonlinear differences in stroke shape. \textbf{Figure 7} shows that the reconstruction of digit "5" using only the top 2 components results in a digit which closely resembles an "8" and a "3", hence reinforcing the above findings.

Digits "4", "7" and "9" have the most prominent overlap, and are practically indistinguishable just using PC1 and PC2, because they have very similar overall structure (these digits have similar boundary definition and thickness in strokes), along with very similar slant points. \textbf{Figure 7} shows that the reconstruction of digit "7" using only the top 2 components results in a digit which closely resembles a "9", hence reinforcing the above findings.

Thus, while PCA reveals broad structural trends, it does not fully separate digit classes due to its linear nature.

\paragraph{Reconstruction.}
\begin{figure}[H]
    \centering
    \includegraphics[width=1.0\textwidth]{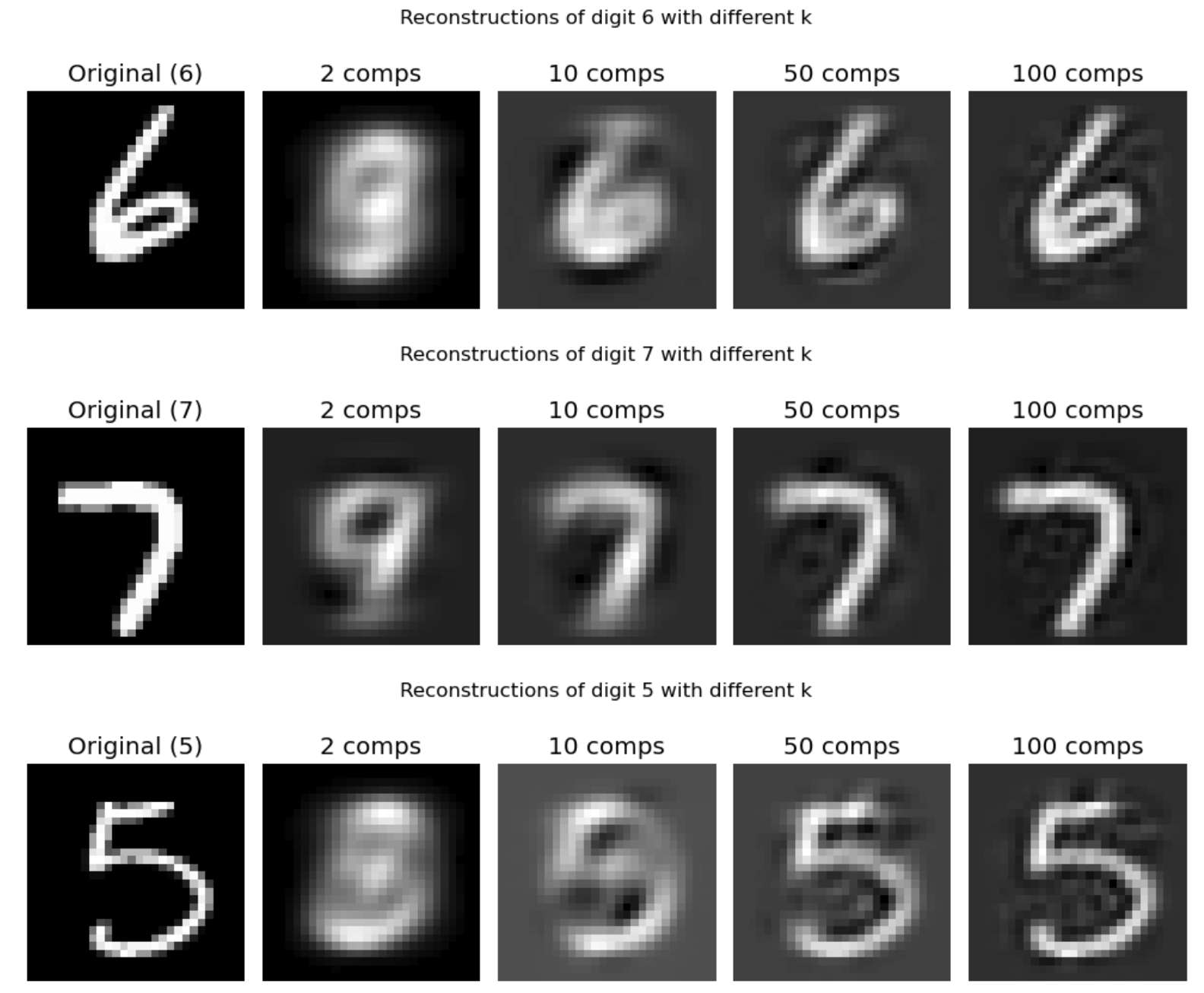}
    \caption{Digit reconstructions using varying numbers of PCA components.}
\end{figure}

Based on the examples in \textbf{Figure 7}, we conclude that reconstructions using the top 50 or more principal components retain nearly all visual information and provide a close approximation of the actual digit which lies in a much higher dimensional (784) plane. 

\subsection{Factor Analysis (FA)}

\subsubsection*{Methodology}

We applied Factor Analysis with 7 latent factors using maximum-likelihood estimation.  
Unlike PCA, which identifies orthogonal directions of total variance, FA decomposes each image as

\[
X = Lz + \varepsilon,
\]

where $L$ contains the factor loadings, $z \sim N(0, I)$ are latent handwriting factors,  
and $\varepsilon \sim N(0, \Psi)$ represents pixel-specific noise (unique variance).  

FA is particularly suited for image data because it recovers shared structure across digits by modeling only common variance and identifies latent handwriting primitives such as loops and diagonal strokes. It complements PCA by separating noise ($\Psi$) from meaningful structure.

\subsubsection*{Results}

\paragraph{Latent Factor Components.}

\begin{figure}[H]
    \centering
    \includegraphics[width=\textwidth]{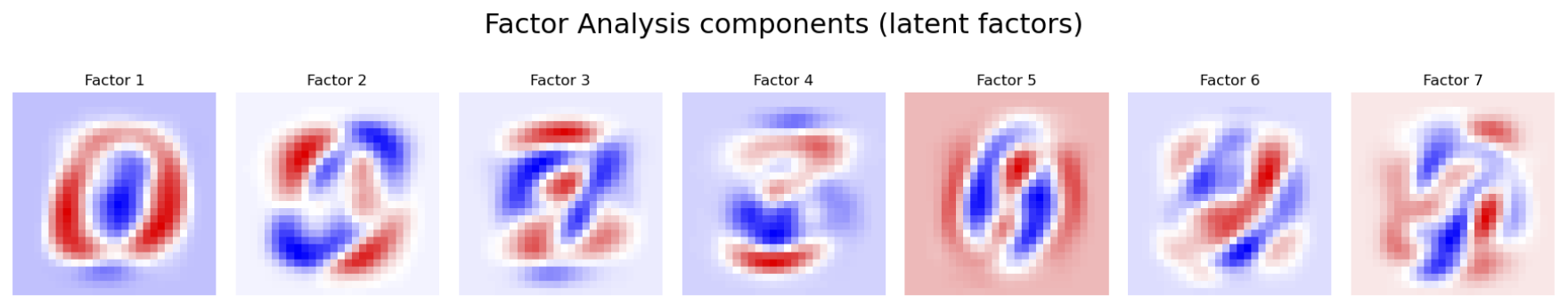}
    \caption{FA latent factors (unrotated loadings).}
\end{figure}

\begin{figure}[H]
    \centering
    \includegraphics[width=\textwidth]{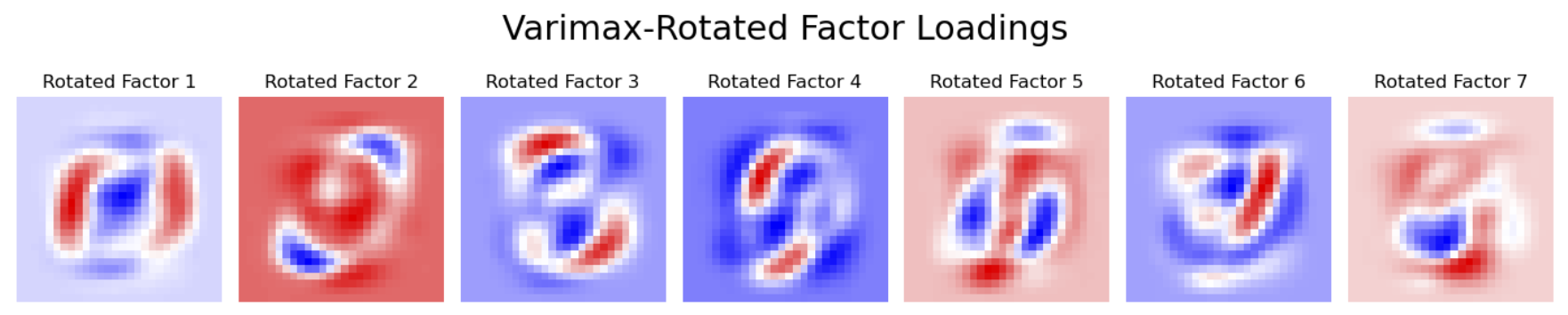}
    \caption{Varimax-rotated factor loadings}
\end{figure}

The unrotated FA loadings (\textbf{Figure 8}) already reveal broad structural trends, but the varimax rotation (\textbf{Figure 9}) substantially improves interpretability by producing factors that correspond to clear handwriting primitives. Factor 1 shows a strong left–right contrast with a central vertical bar. Factor 2 has a ring-like structure suggestive of circular strokes. Factor 3 and 4 concentrate curvature in top-left to bottom-right diagonal. Factor 5 emphasizes lower-half curvature. Factor 6 focuses on right-slant vertical structure and Factor 7 shows an S-shaped curvature. Together, these rotated components provide a clean decomposition of MNIST handwriting variation.

These rotated factors demonstrate that FA decomposes MNIST digits into interpretable handwriting primitives like loops, slants, bars, and symmetric curvature.  
Compared to PCA, whose components mix multiple structural elements, FA provides a  
stroke-level decomposition of digit formation.

\paragraph{FA Embedding.}

\begin{figure}[H]
    \centering
    \includegraphics[width=0.7\textwidth]{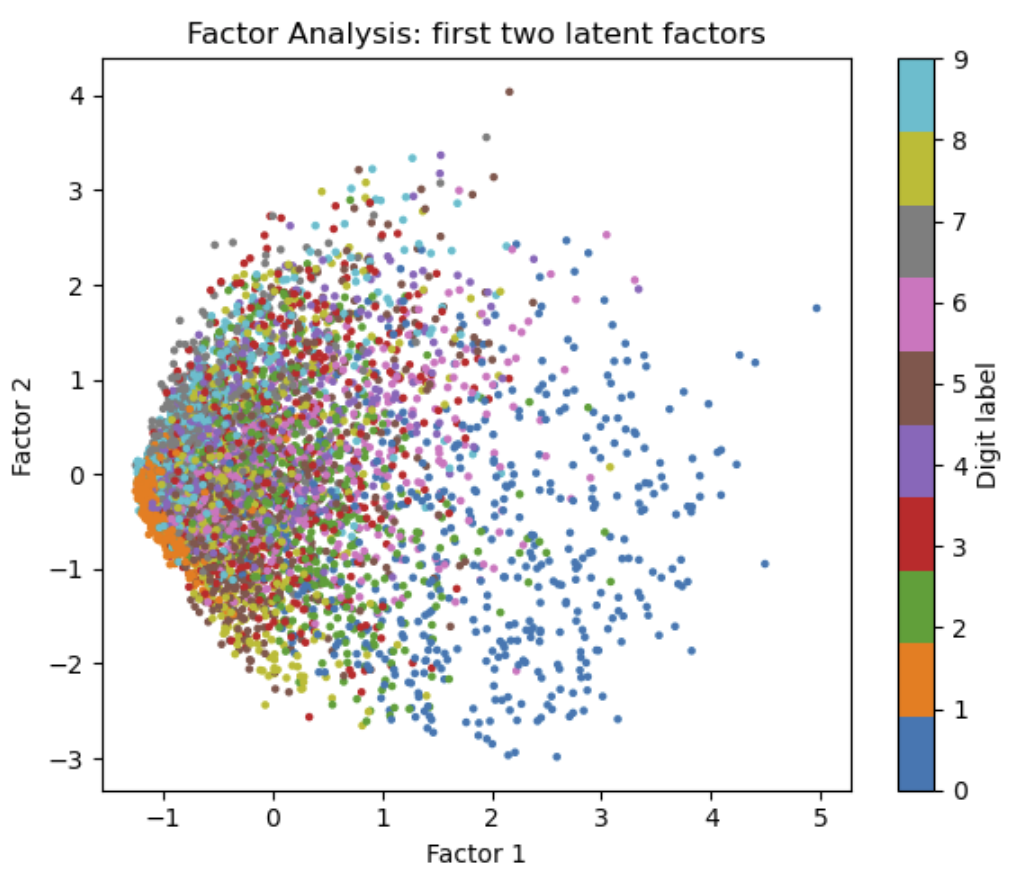}
    \caption{Projection onto the first two FA factors.}
\end{figure}

The 2D embedding (\textbf{Figure 10}) shows a fanning pattern with heavy overlap.  
Digits sharing similar primitive strokes (e.g., loops, diagonals) overlap.

Digit ``0'' forms a mild cluster along high values of Factor~1, reflecting that loop
structure captured by Factor~1 is strongest for it. Digit ``1'' tends to cluster near
small, negative values of both factors because it naturally lacks loops, diagonals, or
curves.

Digits ``3'', ``5'', and ``8'' heavily overlap due to shared curved and S-shaped
components.

Digits ``4'', ``7'' and ``9'' remain difficult to separate because they share diagonal and boundary structure.

 PCA focuses on directions of maximum total variance, often separating digits better, while FA captures only shared variance across digits, so embeddings reflect common handwriting structure rather than digit identity. Thus, FA provides complementary insight to PCA.

\subsection{Uniform Manifold Approximation and Projection (UMAP)}

\subsubsection*{Methodology}

We used UMAP to visualize the nonlinear geometry of handwritten digit images. Unlike PCA, which is restricted to linear projections, UMAP preserves local neighborhoods
and approximates the underlying low-dimensional manifold on which MNIST digits lie. We experimented with several values of \texttt{n\_neighbors}. The final embedding uses $n\_neighbors = 15$, which provided the best balance between local detail and global structure.

\subsubsection*{Results}

\begin{figure}[H]
\centering
\includegraphics[width=0.6\textwidth]{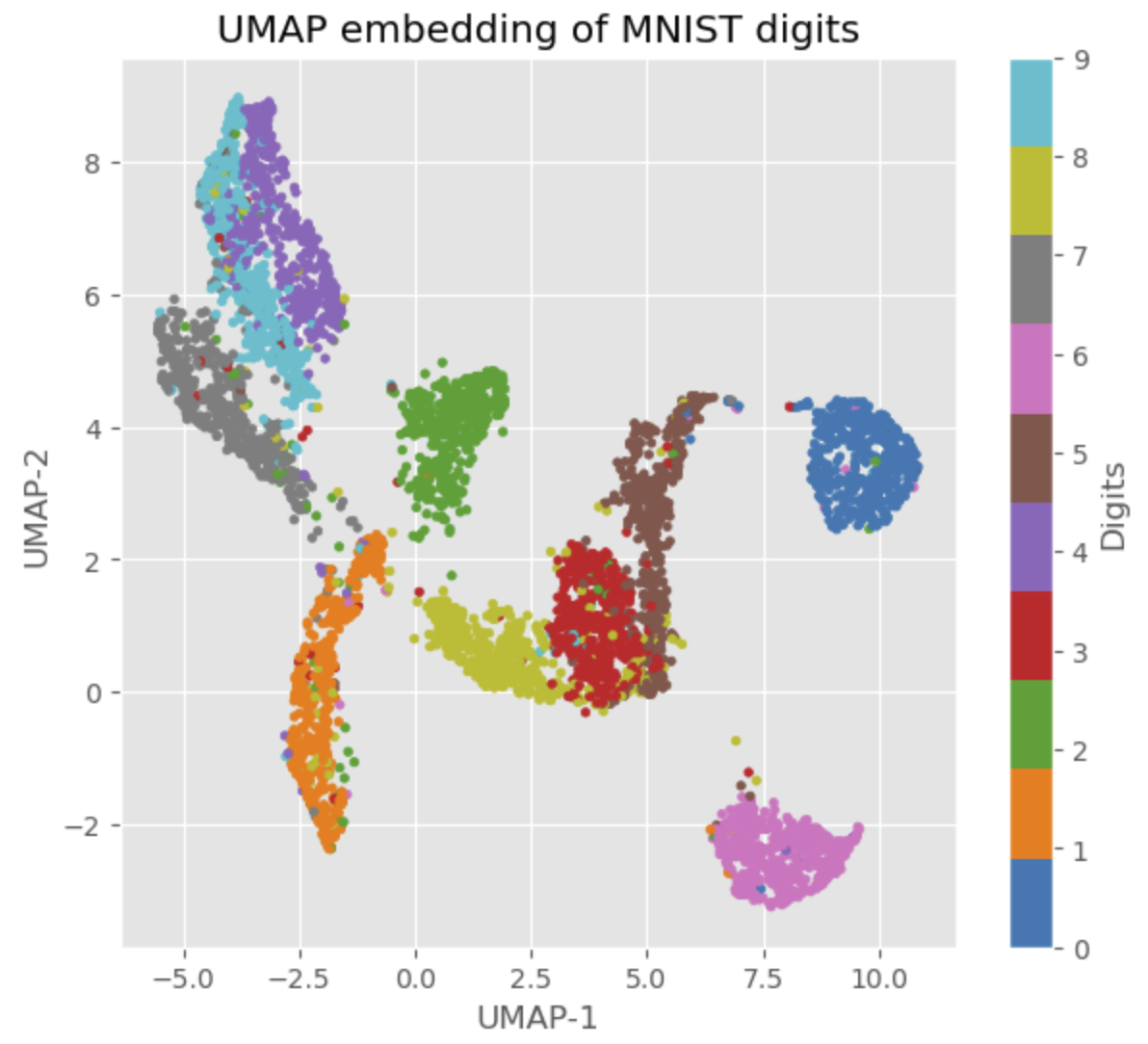}
\caption{UMAP embedding using $n\_neighbors=15$.}
\end{figure}

\paragraph{2D UMAP Clusters}

The two-dimensional UMAP embedding in \textbf{Figure 11} reveals several geometric patterns across the digit classes. Digit "1" appears as a compact and isolated cluster, reflecting its nearly one-dimensional structure as a single vertical stroke. Digit "0" forms a distinctive circular cluster, which captures its closed-loop topology in a way that linear methods such as PCA cannot. Digits "3", "5", and "8" still tend to overlap because they share curved and S-shaped features, and UMAP arranges them along a continuous manifold of curvature styles, still separating the non-linearity much better than PCA and FA. Digits "4", "7", and "9" are also clustered close by, with "9" and "4" having the highest overlap. This could be driven by their common diagonal strokes and similar boundary structure. Digit "6" lies on a curved embedding as well, well-separated from other digits, which reflects the distinct loop in the structure of "6". Finally, the embedding of digit "2" is along an elongated curve, capturing natural variation in its slant structure. Overall, UMAP highlights nonlinear relationships between digits and reveals meaningful structural similarities that are not as apparent under linear projections. Where PCA flattens variations into straight-line directions, UMAP reveals smooth curves, arcs, and rings corresponding to natural handwriting variation.

\begin{figure}[H]
\centering
\includegraphics[width=\textwidth]{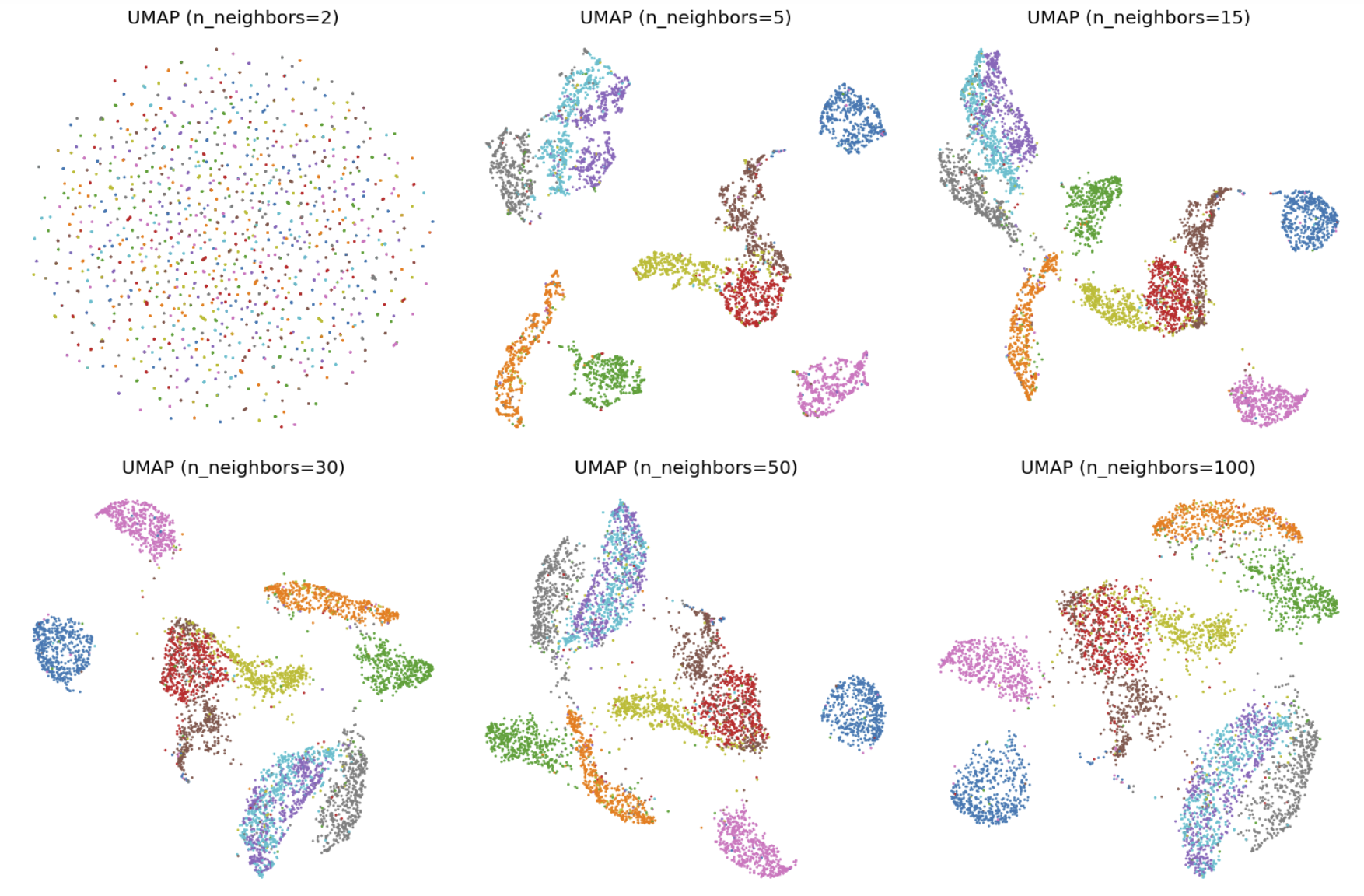}
\caption{UMAP embeddings under varying values of $n\_neighbors$.}
\end{figure}

Varying \texttt{n\_neighbors} in \textbf{Figure 12} reveals how UMAP trades off local detail and global structure. With $n\_neighbors$ = 2, the embedding almost completely falls apart: points scatter into a loose circular cloud with no identifiable digit clusters, as UMAP focuses entirely on extremely local relationships. At $n\_neighbors$ = 5, meaningful digit-wise groupings begin to appear, but the layout is still broken into many small, irregular fragments, reflecting sensitivity to fine stylistic differences rather than overall digit geometry. By $n\_neighbors$ = 15, the embedding becomes much more coherent; digits form stable, well-separated clusters, and the global arrangement resembles the MNIST manifold. Increasing to $n\_neighbors$ = 30 produces clusters that remain distinct but begin to stretch along smooth curves, emphasizing broader manifold contours rather than fine distinctions. At $n\_neighbors$ = 50, several clusters elongate further and draw closer together, revealing higher-level similarities between digits with shared curvature or stroke direction. Finally, at $n\_neighbors$ = 100, the embedding becomes the most globally organized: clusters partially merge or blend at their boundaries, and the layout begins to resemble a continuous manifold rather than sharply separated groups, mirroring the behavior of PCA while still preserving nonlinear structure.

This sensitivity analysis demonstrates that UMAP provides a flexible framework for exploring 
both micro-level handwriting variations and macro-level digit relationships.

To conclude, UMAP offers the most expressive geometric representation among the three methods studied since PCA cannot reveal nonlinear structure while FA embeddings show heavy overlap. UMAP uncovers curved, looped, and slanted manifolds reflecting true handwriting variability. 
\section{Discussion}

This study examines the MNIST dataset through three complementary approaches - PCA,
Factor Analysis, and UMAP - to understand how different dimensionality reduction methods
reveal structure in high-dimensional image data. Rather than focusing on classification, the goal was to study how each method characterizes variation, dimensionality, and geometry in a dataset where visual interpretation is possible.

Across all analyses, a consistent theme emerges: handwritten digits lie on a structured, low-dimensional manifold, but different methods expose different aspects of this structure. PCA provides a variance-based summary that highlights how digits differ globally, especially in overall stroke thickness and symmetry. FA, on the other hand, decomposes digits into interpretable building blocks such as loops and strokes. These latent factors are less about distinguishing digits and more about identifying common handwriting primitives that recur across multiple classes. UMAP offers yet another perspective by uncovering curved and nonlinear manifolds along which handwriting styles vary smoothly.

A notable observation is that these methods do not contradict each other, rather, each reveals a different and valid form of structure. PCA clarifies which global directions dominate variation, FA exposes what structural components cause that variation, and UMAP shows how digits organize themselves when geometric relationships are allowed to be nonlinear. Together, these perspectives suggest that MNIST combines both discrete class boundaries and continuous stylistic transitions.

\subsection*{Limitations}

Several limitations should be acknowledged.  
First, both PCA and FA are linear and can only capture additive combinations of pixel values. As a result, nonlinear transitions such as the curvature differences that distinguish some digits cannot be fully represented in their low-dimensional embeddings. FA also requires rotational adjustments to achieve interpretability, and the resulting factors, though meaningful, are not unique.  
Second, UMAP is sensitive to hyperparameters and randomness, and while visually compelling, its embeddings do not correspond to a probabilistic model or a reconstruction framework. Finally, because this study focused on exploratory structure rather than prediction, the practical implications for downstream tasks such as digit recognition and prediction were not investigated.

\subsection*{Potential Future Work}

Future work could extend this analysis in several ways.  
Applying nonlinear extensions such as kernel PCA or diffusion maps would allow direct comparison with UMAP under a more formal eigen-decomposition framework. A Bayesian or sparse version of FA could yield sharper and more interpretable factors by encouraging localized loadings. It would also be valuable to evaluate how well low-dimensional embeddings support classification, clustering, or anomaly detection, thereby linking geometric insights with practical performance. Finally, CCA could also be explored to quantify correlations between different spatial regions of digits, offering a view on any symmetry in handwriting styles.

Overall, the combined results demonstrate that no single method fully captures the complexity of handwritten digit structure. However, taken together, PCA, FA, and UMAP provide a rich and multifaceted understanding of MNIST, one that blends global variance structure, latent stroke composition, and nonlinear geometric organization.

\section{References}

\begin{itemize}

    \item E. Tuzhilina, Lecture notes on multivariate analysis.
    \item P. Zwiernik, \emph{Lecture notes on multivariate analysis (STA437)}.  
Available at \url{https://pzwiernik.github.io/sta437/}  

\item P. Zwiernik, \emph{Lecture notes on principal component analysis (STA414)}.  
Available at \url{https://pzwiernik.github.io/sta414/}

\item R. G. Krishnan, \emph{Lecture notes on statistical and machine learning methods (CSC311)}.  
Available at \url{https://www.cs.toronto.edu/~rahulgk/courses/csc311_f24/index.html}

    \item scikit-learn documentation for PCA and Factor Analysis:

    \begin{itemize}
        \item \url{https://scikit-learn.org/stable/modules/generated/sklearn.decomposition.PCA.html}    
        \item \url{https://scikit-learn.org/stable/modules/generated/sklearn.decomposition.FactorAnalysis.html}
    
        \item \url{https://scikit-learn.org/stable/auto_examples/decomposition/plot_varimax_fa.html}
    \end{itemize}

    \item Documentation for UMAP:
    \begin{itemize}

    \item \url{https://umap-learn.readthedocs.io/en/latest/basic_usage.html}
    
    \end{itemize}
    
    \item Writing assistance tools were used for language editing and LaTeX formatting.

\end{itemize}

\end{document}